\documentclass{article}

\usepackage[preprint, nonatbib]{neurips_2021}

\usepackage[utf8]{inputenc} 
\usepackage[T1]{fontenc}    
\usepackage[colorlinks]{hyperref}       
\usepackage{url}            
\usepackage{booktabs}       
\usepackage{amsfonts}       
\usepackage{nicefrac}       
\usepackage{microtype}      
\usepackage{xcolor}         

\usepackage{tikzsymbols}
\usetikzlibrary{arrows}
\usetikzlibrary{chains}
\usetikzlibrary{patterns}
\usepackage{tikz-qtree}
\usetikzlibrary{shapes.multipart}
\usetikzlibrary{positioning}
\usetikzlibrary{arrows.meta}
\usepackage{pgfplots}

\usetikzlibrary{calc}
\usetikzlibrary{positioning}
\usepgflibrary{fpu}
\usepgfplotslibrary{fillbetween}

\pgfplotsset{compat=newest}

\usepackage{amsmath}
\usepackage{amssymb}
\usepackage{placeins}
\usepackage{wrapfig}
\usepackage{subcaption}

\title{Feature Disentanglement of Robot Trajectories}

\author{%
    \parbox{\linewidth}{
        \centering
        Matias Valdenegro-Toro$^1$ \quad Daniel Harnack$^{1}$ \quad Hendrik Wöhrle$^{1}$\\
    }\\
    ~\\
    \parbox{\linewidth}{
        \centering
        $^1$ German Research Center for Artificial Intelligence, 28359 Bremen, Germany.\\
        ~\\
        \texttt{matias.valdenegro@dfki.de}, \texttt{daniel.harnack@dfki.de}, \texttt{hendrik.woehrle@dfki.de}
    }
}

\begin{document}

\maketitle

\begin{abstract}
    Modeling trajectories generated by robot joints is complex and required for high level activities like trajectory generation, clustering, and classification. Disentagled representation learning promises advances in unsupervised learning, but they have not been evaluated in robot-generated trajectories. In this paper we evaluate three disentangling VAEs ($\beta$-VAE, Decorr VAE, and a new $\beta$-Decorr VAE) on a dataset of 1M robot trajectories generated from a 3 DoF robot arm. We find that the decorrelation-based formulations perform the best in terms of disentangling metrics, trajectory quality, and correlation with ground truth latent features. We expect that these results increase the use of unsupervised learning in robot control.
\end{abstract}

\section{Introduction}

Robots need to understand and learn from their environment. A large part of the literature is devoted to perception and control in predefined environments, but generalization and learning should happen in completely new and unseen environments. Trajectory modeling is one part of robot control and planning that generally has issues with generalization across new environments \cite{toussaint2009robot} \cite{fabisch2020learning}.

Unsupervised learning is a common approach for this problem, but it makes assumptions on data. Applying unsupervised learning to trajectory generation and control would allow for a Robot to improve its generalization and to learn from the environment without human supervision, achieving the goals of long-term autonomy.

Modeling of robots trajectories in joint space is complicated due to non-linear behavior, singularities, and discretization required to gather training data \cite{fabisch2020learning}. Learning features from trajectories in order to build new trajectories would be useful as a general tool for trajectory modeling \cite{toussaint2009robot}, and would enable other robot features such as generating new trajectories, decomposing them into simpler movements \cite{gutzeit2021comparison}, classifying trajectories into semantic movements, clustering trajectories, and identifying appropriate trajectories for a given task \cite{natarajan2021}.

Variational Autoencoders (VAEs) are a staple technique to perform unsupervised learning. But the features they learn as generative models are often correlated. Disentangling VAEs are used for the purpose of learning decorrelated (aka disentangled) features, with varying success. These kind of VAEs are often evaluated with image datasets, and there is little research using them with trajectories and paths.

One of the first disentangling formulations is the $\beta$-VAE \cite{higgins2016beta}, where the $\beta$ parameter controls the disentangling level by increasing the weight of the KL divergence with the standard normal distribution (which is decorrelated). Another disentangling method is to use a decorrelated prior or to induce decorrelation in the loss formulation \cite{locatello2019challenging}. There are some formulations available for sequence data like \cite{yingzhen2018disentangled} and \cite{hsu2017unsupervised} but these are too complex and make unrealistic assumptions (highly dimensional, weak temporal correlations) for robot trajectory data.

In this paper we evaluate three disentangling VAE formulations for the purpose of learning disentangled features from trajectories of a Robot arm.

The contributions of this paper are: A dataset of robot trajectories for evaluation of disentangling methods, including ground truth latent features. An evaluation of three disentangling VAE formulations, and validation on how these methods perform in modeling robot joint trajectories.

\begin{figure}[t]
    \centering
    \includegraphics[width=\textwidth]{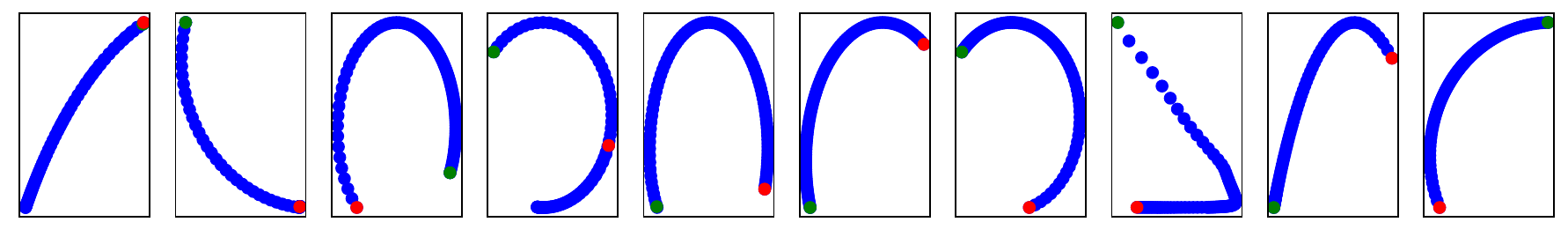}
    \includegraphics[width=\textwidth]{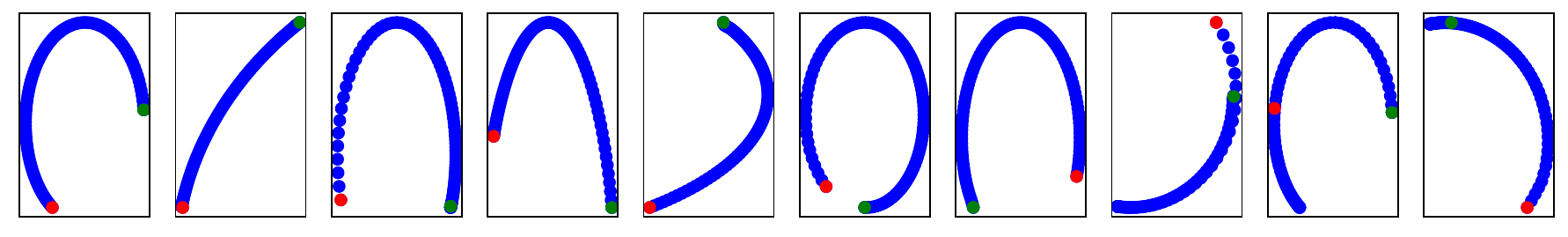}
    \caption{A random sample of ten trajectories from the training (top row) and testing (bottom row) sets. Note that the scale has been omitted for visualization purposes.  The green point is the trajectory start, and the red one is its end.}
    \label{trajectory_samples}
\end{figure}

\section{Trajectory Feature Disentanglement}

The purpose of feature learning is to generalize to multiple downstream tasks. But these features can often be correlated, decreasing their power. Learning decorrelated features is generally called disentangling in the literature. In this section we describe our variational autoencoder (VAE) formulation that we use to learn disentangled features.

We use the following general loss formulation $L(x, \hat{x}, z) = \text{MSE}(x, \hat{x}) -\beta \text{KL}(z, \mathrm{N}(0, 1)) + \gamma \text{COV}(z)$.
With $x$ being the input trajectory, $\hat{x}$ is the predicted trajectory, and $z$ are the latent features learned by the VAE. The COV term is given by $\text{COV}(z) = 0.5\left[\sum_i^F \text{cov}(z)_i^2 - \sum_i \text{diag}(\text{cov}(z))_i^2 \right]$. 

Where $i$ indexes feature dimensions and $\text{cov}(z)$ computes covariance matrix of features $z$. Additional terms of the loss are $\text{COV}(z)$ as the decorrelation loss \cite{cogswell2015reducing} that penalizes for cross-elements in the covariance matrix of the latent features $z$, $\text{KL}$ is the Kullback-Leibler Divergence, used as a distance between probability distributions, and $\text{MSE}$ is the mean squared error used as reconstruction error during training.

If $\beta = 0$ and $\gamma = 0$, this is the standard autoencoder formulation, while if $\beta = 1 $ and $\gamma = 0$, this is the standard variational autoencoder formulation, and $\beta > 1$ and $\gamma = 0$ defines the $\beta$-VAE \cite{higgins2016beta} disentangling formulation, and $\beta = 1$ and $\gamma > 0$ is the decorrelation disentangling autoencoder formulation \cite{chen2018isolating}.

We also build a combination of the $\beta$ and decorrelation VAE which we call $\beta$-decorr VAE, where $\beta > 1$ and $\gamma > 0$. This formulation performs disentangling by imposing a penalty with the KL between the standard normal distribution and the distribution of the latent features $z$, with additional penalty given by the decorrelation loss applied to $z$.

\section{Experimental Evaluation}

In this section we evaluate different methods for disentangled feature learning and our proposed method $\beta$-decorr VAE.

\begin{wrapfigure}{r}{0.3\textwidth}
    \includegraphics[width=\linewidth]{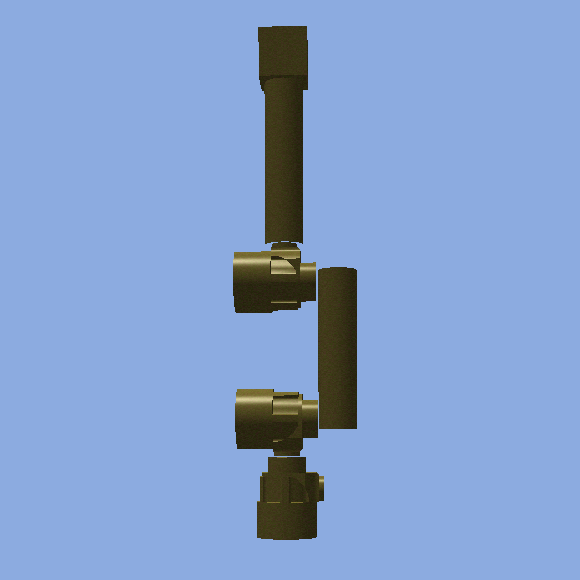}
    \caption{The 3-DoF robot arm used to generate trajectories in our dataset. Joints are base pan-tilt (2 DoF) and a 1 DoF joint between links.}
    \label{arm3dof}
\end{wrapfigure}

\textbf{Dataset}. We use a dataset of robot trajectories generated using a power series representation of the input to the joints. The robot is a simple 3-DoF manipulator (shown in Figure \ref{arm3dof}). We generated 1 million trajectories using a series of random movements, and the power series features are stored as ground truth features. Two kinds of trajectories are available, 2D (XY) for simplicity, and 3D (XYZ). Each trajectory is 100 timesteps long, with 5 dimensions per point (XY and joint positions) for the 2D case, and 6 dimensions per point (XYZ and joint positions) for the 3D case. Ground truth latent features are $F = 15$ dimensional for both cases.

We perform a 70\% training and 30\% testing split. A random sample of ten trajectories for train and test sets are shown in Figure \ref{trajectory_samples}. For more details on the dataset, please see the supplementary material.

\textbf{Metrics}. We use several metrics to assess different aspects of performance. The reconstruction mean absolute error (MAE) measures how well trajectories are reconstructed. For disentangling performance, we use the KL (lower is better) between the latent feature distribution and a 15-dimensional standard normal distribution with diagonal covariance. We also measure the pearson correlation between the predicted and ground truth latent features and build correlation matrices between all feature pairs. This measures if the disentangling VAEs are recovering the true latent features or if they learn a different set of features.

In the supplementary we make multiple plots comparing pairwise metrics, such as MAE vs KL, or KL vs mean correlation, which show different trade-offs in disentangling performance. It must be mentioned that there is a common trade-off between disentangling performance and reconstruction error, better disentangling usually entails worse reconstruction and viceversa \cite{locatello2019challenging}.

\textbf{Hyper-parameters}. For $\beta$-VAE, we use $\beta \in [1.0, 5.0, 10.0, 50.0, 100.0]$, for decorr-VAE we use $\gamma \in [0.0, 0.01, 0.1, 1.0, 5.0, 10.0, 50.0, 100.0]$, and for $\beta$-decorr VAE we use the cross product of the two previous values $\beta, \gamma = [\beta_{\beta-\text{VAE}} \times \gamma_{\text{decorr}}]$ as both a $\beta$ and $\gamma$ values are needed to train one VAE instance.

\textbf{Results}. Our main results are presented in Table \ref{disentangle_comparison}, with plots presented in Figures \ref{plots_betavae2D}, \ref{plots_decorrvae2D}, and \ref{plots_betadecorrvae2D} (in the supplementary). Section \ref{trajectories_sec} of the supplementary contains a selection of scatter plots of the generated trajectories through disentangled features.

\begin{table}[t]
    \centering
    \begin{tabular}{llllll}
        \toprule
        Model	& $\beta$	& $\gamma$	& Reconstruction MAE	& Latent KL	& Mean Corr $\mu_c$\\
        \toprule
        AE		& 0			& 0			& 0.022					& NA		& 0.64\\
        VAE 	& 1			& 0			& 0.097					& 30.09		& 0.39\\
        \midrule
        \multicolumn{6}{l}{Criteria - Best Mean Correlation}\\
        \midrule        
        $\beta$-VAE & 5 	& 0	& 0.183		& 17.26		& 0.58\\
        Decorr VAE 	& 1		& 5	& 0.110		& 29.17		& 0.60\\
        $\beta$-Decorr VAE	& 5 & 5 	& 0.182	& 17.47	& \textbf{0.64}\\
        \midrule
        \multicolumn{6}{l}{Criteria - Best Latent KL Disentangling} \\
        \midrule
        $\beta$-VAE & 100	& 0	& 0.614			& 2.48		& 0.49\\
        Decorr VAE 	& 1		& 0.1 & 0.109		& 28.73	& 0.52\\
        $\beta$-Decorr VAE	& 100 & 10 	& 0.615	& \textbf{2.42}	& 0.53\\
        \midrule
        \multicolumn{6}{l}{Criteria - Best Reconstruction Error} \\
        \midrule
        $\beta$-VAE & 1		& 0	& 0.097			& 30.09		& 0.39\\
        Decorr VAE 	& 1		& 100 & 0.086		& 36.82		& 0.54\\
        $\beta$-Decorr VAE	& 0.1 & 5 	& \textbf{0.041}	& 55.76		& 0.52\\
        \bottomrule
    \end{tabular}
    \caption{Comparison of varying disentangling autoencoders with different metrics.}
    \label{disentangle_comparison}
\end{table}

In Table \ref{disentangle_comparison}, we use three criteria to select the best method based on the hyper-parameters, namely mean correlation with ground truth features, best disentangled features (as measured by the latent KL), and best reconstruction error. As baselines we also present the metrics and trajectories produced by a standard autoencoder (AE) and a VAE (corresponding to $\beta$-VAE with $\beta = 1.0$). Note that a standard AE does not perform useful feature learning as the feature distribution is basically unknown.

In terms of all metrics, the $\beta$-Decorr VAE outperforms all other methods. This indicates that the effect of the KL penalty and the decorrelation applied to the latent features interact in order to provide additional disentangling than just the KL penalty or decorrelation individually.

Looking at individual production of trajectories given single features (Sec \ref{trajectories_sec}), the highest quality trajectories are produced by a VAE, and the trajectory quality degrades with disentangled features (as commonly happens in other modalities \cite{locatello2019challenging}). $\beta$-VAE produces noisy trajectories which was unexpected, and we believe this might be due to unstable training dynamics produced by the KL penalty.

\begin{figure}[t]
    \centering
    \begin{subfigure}{0.297\textwidth}
        \includegraphics[width=0.99\linewidth]{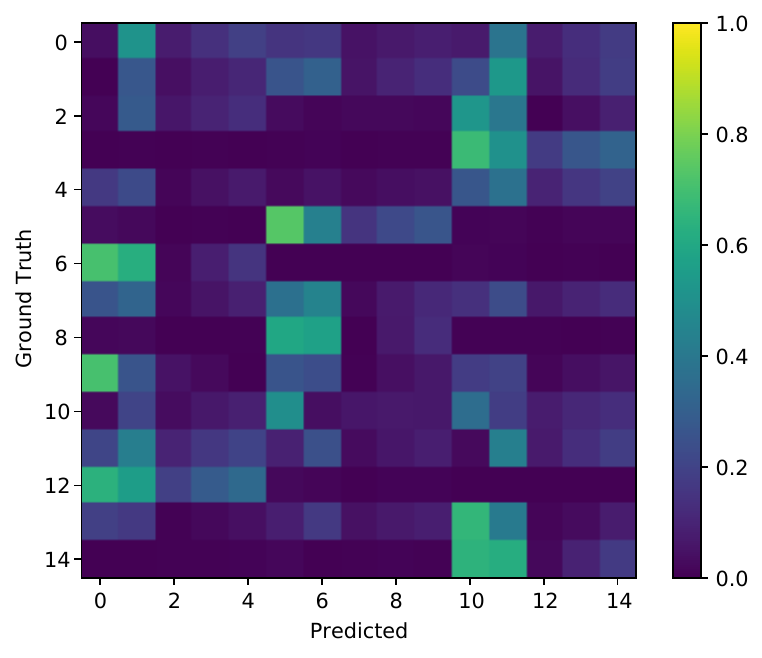}
        \caption{$\beta$-VAE\\($\beta = 5.0$, $\mu_c = 0.58$)}
    \end{subfigure}
    \begin{subfigure}{0.2565\textwidth}
        \includegraphics[width=0.99\linewidth]{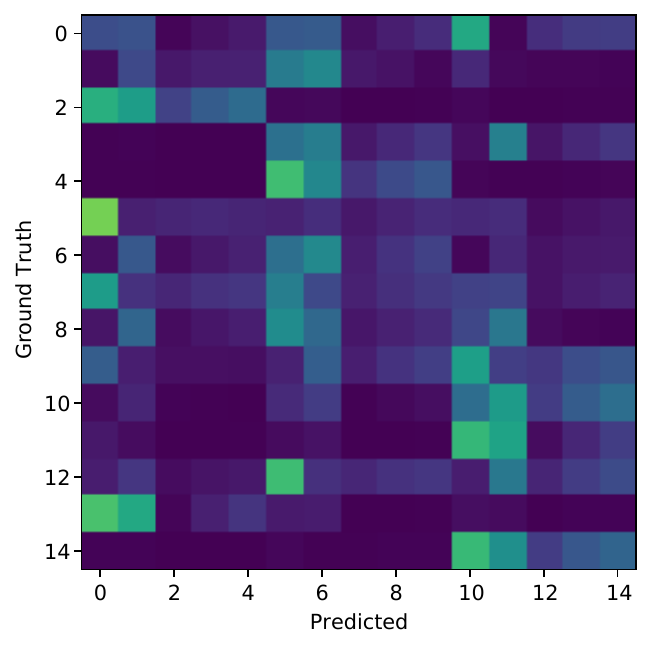}
        \caption{Decorr-VAE\\($\gamma = 5.0$, $\mu_c = 0.60$)}
    \end{subfigure}
    \begin{subfigure}{0.2565\textwidth}
        \includegraphics[width=0.99\linewidth]{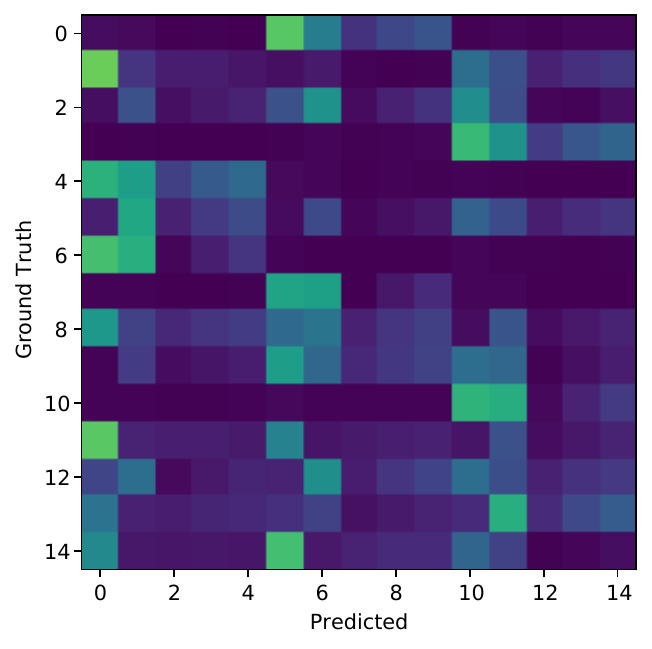}
        \caption{$\beta$-Decorr VAE ($\beta = 5.0 \\ \gamma = 5.0$, $\mu_c = 0.64$)}
    \end{subfigure}
    \caption{Correlation matrices between predicted and ground truth features produced by models according to the best mean correlation criteria, which indicates the level of true features that can be recovered. Each heatmap has the same colorbar scale.}
    \label{corr_matrices}
\end{figure}

Decorr VAE produces mostly smooth trajectories when compared to $\beta$-VAE. We believe this shows that the decorrelation loss has a lower effect on the reconstructed trajectories, specially for higher values of the $\gamma$ parameters. The trajectories produced by $\beta$-Decorr VAE allows comparing the effect of $\beta$ and $\gamma$ parameters. Looking at Figure \ref{plots_betadecorrvae2D}, increasing $\beta$ immediately means that the reconstruction error as measured by the MAE increases, but with $\gamma$ there seems to be no direct relationship, with the value of MAE only varying with $\beta$. This supports our conclusion that the decorrelation formulation of a disentangling VAE is superior to the $\beta$-VAE formulation.

Overall the best mean correlation with ground truth features is 0.64, obtained by the standard autoencoder and $\beta$-Decorr VAE, which is quite low as it explains only $0.64^2 = 0.41 = 41\%$ of the feature variance. Figure \ref{corr_matrices} shows the correlation metrices for models selected by best mean correlation, and it can be visually seen how features are correlated. All methods show different levels of multiple correlations with the true features (lines in the correlation matrix), while $\beta$-Decorr VAE shows slightly smaller levels of correlation, with stronger peaks for individual true features.

\section{Conclusions and Future Work}

In this paper we evaluate three disentangling VAE formulations for the task of learning disentangled features of robot trajectories. We find that the best performing formulations are the ones using a decorrelation loss to penalize non-diagonal correlations between latent features, which seems to work better in terms of learning stability and quality of output trajectories than other formulations such as the $\beta$-VAE.

We expect that our results will guide the development and future use of unsupervised learning techniques in robots, particularly for control, like trajectory generation, path planning, or trajectory reuse for multiple downstream tasks.

In our paper we mostly analyze the trajectories generated by the VAE decoder, but the disentangled features produced by the encoder are also useful, for example for movement classification \cite{gutzeit2021comparison}, fault detection, and clustering of trajectories.

Our dataset will be make publicly available and we hope it will be useful to the community.

\FloatBarrier
\clearpage

\bibliographystyle{plain}
\bibliography{biblio}

\clearpage
\appendix

\section{Robot Trajectory Dataset}

We use the following formulation to generate trajectories using a polynomial input representation.

\begin{equation}
    q(\phi) = -q_0 (\phi - 1) + q_1 \phi_1 + \sum_{i=2}^{N_q} q_i(\phi^{i-1} - 1) \phi
\end{equation}

Where $\phi \in [0, 1]$ is the joint phase and $N_q$ is the number of parameters for each joint. $q_0$ is the joint start and $q_1$ is the joint end positions. We use $N_q = 5$, corresponding to five polynomial parameters for each joint ($F = 15$ latent features in total), and sample $q_i$ values from $[-\pi, \pi]$ as this is the range of each joint, which naturally produces a good selection of trajectories.

We generate 1 Million trajectories using this method, and perform a train/test split with ratios  70\%/30\%, producing a training set of 700 K trajectories, and a test set of 300 K trajectories.

\section{Neural Network Architecture and Training Setup}

We use multi-layer perceptron encoder-decoder architecture. While a recurrent neural network would be a more natural choice, using RNNs or LSTMs produced strong vanishing or exploding gradients due to the long trajectory length (100 timesteps). MLP architectures did not have this problem.

The network architecture is FC(512)-FC(256)-FC(128)-FC(128)-FC(latent dim)-FC(128)-FC(128)-FC(256)-FC(512)-FC(traj dims), where latent dim is 15 for our dataset. All layers except the last and latent dimension layers use a ReLU activation. The last layer uses a linear activation, while the latent dimension layer also uses a linear activation, but contains two output heads, one for the mean and another for the logarithm of the variance of a Gaussian distribution that is sampled to produce a latent feature. Batch normalization is used between all layers.

This network is trained using the Adam optimizer with a learning rate $\alpha = 0.001$ for 100 epochs using a batch size of 256 samples.

\FloatBarrier
\section{Plots and Metrics for Selected Disentangling VAEs}

The following figures show the relationship between different task (reconstruction mean absolute error) and disentangling (KL Latent, Mean Correlation) metrics, and the different parameters ($\beta$, $\gamma$) of the various disentangling VAEs we evaluated. In particular plots comparing KL Latent and MAE show the trade-off between disentangling and reconstruction quality.

\begin{figure}[h]
    \centering
    \begin{tikzpicture}
        \begin{axis}[height = 0.18 \textheight, width = 0.32\linewidth, xlabel={MAE}, ylabel={Mean Corr}, grid style=dashed, legend pos = north east, legend style={font=\scriptsize}, tick label style={font=\scriptsize}]
            
            \addplot+[scatter, only marks, scatter src=y] table[x  = val_mae, y  = mean_corr, col sep = semicolon] {data/search-mlpvae-betavae-2D-norm.csv};
        \end{axis}		
    \end{tikzpicture}
    \begin{tikzpicture}
        \begin{axis}[height = 0.18 \textheight, width = 0.32\linewidth, xlabel={MAE}, ylabel={KL Latent}, grid style=dashed, legend pos = north east, legend style={font=\scriptsize}, tick label style={font=\scriptsize}]
            
            \addplot+[scatter, only marks, scatter src=y] table[x  = val_mae, y  = val_kl_latent, col sep = semicolon] {data/search-mlpvae-betavae-2D-norm.csv};
        \end{axis}		
    \end{tikzpicture}
    \begin{tikzpicture}
        \begin{axis}[height = 0.18 \textheight, width = 0.32\linewidth, xlabel={Mean Corr}, ylabel={KL Latent}, grid style=dashed, legend pos = north east, legend style={font=\scriptsize}, tick label style={font=\scriptsize}]
            
            \addplot+[scatter, only marks, scatter src=y] table[x  = mean_corr, y  = val_kl_latent, col sep = semicolon] {data/search-mlpvae-betavae-2D-norm.csv};
        \end{axis}		
    \end{tikzpicture}
    \begin{tikzpicture}
        \begin{axis}[height = 0.18 \textheight, width = 0.32\linewidth, xlabel={Beta}, ylabel={KL Latent}, grid style=dashed, legend pos = north east, legend style={font=\scriptsize}, tick label style={font=\scriptsize}, xmode=log]
            
            \addplot+[scatter, only marks, scatter src=y] table[x  = beta, y  = val_kl_latent, col sep = semicolon] {data/search-mlpvae-betavae-2D-norm.csv};
        \end{axis}		
    \end{tikzpicture}
    \begin{tikzpicture}
        \begin{axis}[height = 0.18 \textheight, width = 0.32\linewidth, xlabel={Beta}, ylabel={MAE}, grid style=dashed, legend pos = north east, legend style={font=\scriptsize}, tick label style={font=\scriptsize}, xmode=log]
            
            \addplot+[scatter, only marks, scatter src=y] table[x  = beta, y  = val_mae, col sep = semicolon] {data/search-mlpvae-betavae-2D-norm.csv};
        \end{axis}		
    \end{tikzpicture}
    \caption{Comparison of several disentangling and reconstruction metrics for the 2D case with $\beta$-VAE}
    \label{plots_betavae2D}
\end{figure}
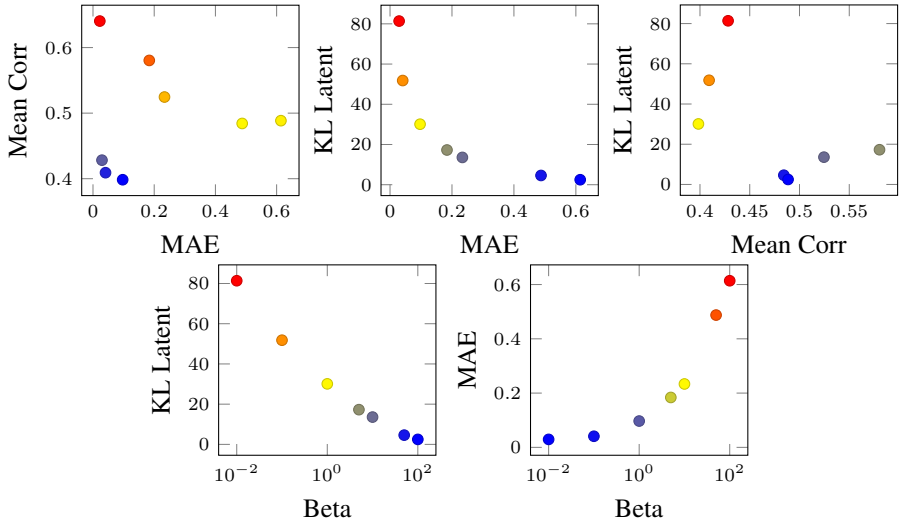

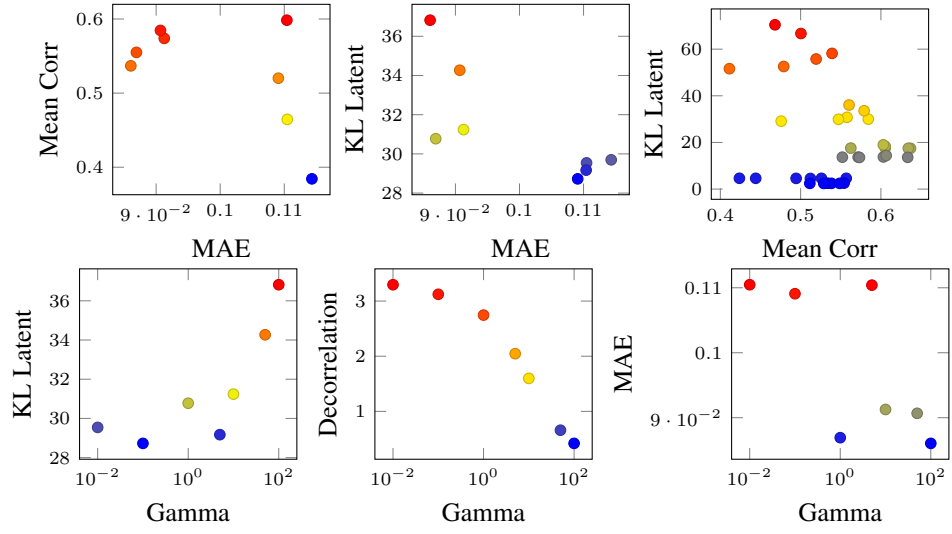
\begin{figure}[h]
    \centering
    \begin{tikzpicture}
        \begin{axis}[height = 0.18 \textheight, width = 0.32\linewidth, xlabel={MAE}, ylabel={Mean Corr}, grid style=dashed, legend pos = north east, legend style={font=\scriptsize}, tick label style={font=\scriptsize}]
            
            \addplot+[scatter, only marks, scatter src=y] table[x  = val_mae, y  = mean_corr, col sep = semicolon] {data/search-mlpvae-decorr-2D-norm.csv};
        \end{axis}		
    \end{tikzpicture}
    \begin{tikzpicture}
        \begin{axis}[height = 0.18 \textheight, width = 0.32\linewidth, xlabel={MAE}, ylabel={KL Latent}, grid style=dashed, legend pos = north east, legend style={font=\scriptsize}, tick label style={font=\scriptsize}]
            
            \addplot+[scatter, only marks, scatter src=y] table[x  = val_mae, y  = val_kl_latent, col sep = semicolon] {data/search-mlpvae-decorr-2D-norm.csv};
        \end{axis}		
    \end{tikzpicture}
    \begin{tikzpicture}
        \begin{axis}[height = 0.18 \textheight, width = 0.32\linewidth, xlabel={Mean Corr}, ylabel={KL Latent}, grid style=dashed, legend pos = north east, legend style={font=\scriptsize}, tick label style={font=\scriptsize}]
            
            \addplot+[scatter, only marks, scatter src=y] table[x  = mean_corr, y  = val_kl_latent, col sep = semicolon] {data/search-mlpvae-decov-2D-norm.csv};
        \end{axis}		
    \end{tikzpicture}
    \begin{tikzpicture}
        \begin{axis}[height = 0.18 \textheight, width = 0.32\linewidth, xlabel={Gamma}, ylabel={KL Latent}, grid style=dashed, legend pos = north east, legend style={font=\scriptsize}, tick label style={font=\scriptsize}, xmode=log]
            
            \addplot+[scatter, only marks, scatter src=y] table[x  = gamma, y  = val_kl_latent, col sep = semicolon] {data/search-mlpvae-decorr-2D-norm.csv};
        \end{axis}		
    \end{tikzpicture}
    \begin{tikzpicture}
        \begin{axis}[height = 0.18 \textheight, width = 0.32\linewidth, xlabel={Gamma}, ylabel={Decorrelation}, grid style=dashed, legend pos = north east, legend style={font=\scriptsize}, tick label style={font=\scriptsize}, xmode=log]
            
            \addplot+[scatter, only marks, scatter src=y] table[x  = gamma, y  = val_decorr, col sep = semicolon] {data/search-mlpvae-decorr-2D-norm.csv};
        \end{axis}		
    \end{tikzpicture}
    \begin{tikzpicture}
        \begin{axis}[height = 0.18 \textheight, width = 0.32\linewidth, xlabel={Gamma}, ylabel={MAE}, grid style=dashed, legend pos = north east, legend style={font=\scriptsize}, tick label style={font=\scriptsize}, xmode=log]
            
            \addplot+[scatter, only marks, scatter src=y] table[x  = gamma, y  = val_mae, col sep = semicolon] {data/search-mlpvae-decorr-2D-norm.csv};
        \end{axis}		
    \end{tikzpicture}
    \caption{Comparison of several disentangling and reconstruction metrics for the 2D case with Decorr VAE}
    \label{plots_decorrvae2D}
\end{figure}

\begin{figure}[!th]
    \centering
    \begin{tikzpicture}
        \begin{axis}[height = 0.18 \textheight, width = 0.32\linewidth, xlabel={MAE}, ylabel={Mean Corr}, grid style=dashed, legend pos = north east, legend style={font=\scriptsize}, tick label style={font=\scriptsize}]
            
            \addplot+[scatter, only marks, scatter src=y] table[x  = val_mae, y  = mean_corr, col sep = semicolon] {data/search-mlpvae-decov-2D-norm.csv};
        \end{axis}		
    \end{tikzpicture}
    \begin{tikzpicture}
        \begin{axis}[height = 0.18 \textheight, width = 0.32\linewidth, xlabel={MAE}, ylabel={KL Latent}, grid style=dashed, legend pos = north east, legend style={font=\scriptsize}, tick label style={font=\scriptsize}]
            
            \addplot+[scatter, only marks, scatter src=y] table[x  = val_mae, y  = val_kl_latent, col sep = semicolon] {data/search-mlpvae-decov-2D-norm.csv};
        \end{axis}		
    \end{tikzpicture}
    \begin{tikzpicture}
        \begin{axis}[height = 0.18 \textheight, width = 0.32\linewidth, xlabel={Mean Corr}, ylabel={KL Latent}, grid style=dashed, legend pos = north east, legend style={font=\scriptsize}, tick label style={font=\scriptsize}]
            
            \addplot+[scatter, only marks, scatter src=y] table[x  = mean_corr, y  = val_kl_latent, col sep = semicolon] {data/search-mlpvae-decov-2D-norm.csv};
        \end{axis}		
    \end{tikzpicture}

    \begin{tikzpicture}
        \begin{axis}[height = 0.18 \textheight, width = 0.32\linewidth, xlabel={Beta}, ylabel={KL Latent}, grid style=dashed, legend pos = north east, legend style={font=\scriptsize}, tick label style={font=\scriptsize}, xmode=log]
            
            \addplot+[scatter, only marks, scatter src=y] table[x  = beta, y  = val_kl_latent, col sep = semicolon] {data/search-mlpvae-decov-2D-norm.csv};
        \end{axis}		
    \end{tikzpicture}
    \begin{tikzpicture}
        \begin{axis}[height = 0.18 \textheight, width = 0.32\linewidth, xlabel={Beta}, ylabel={Decorrelation}, grid style=dashed, legend pos = north east, legend style={font=\scriptsize}, tick label style={font=\scriptsize}, xmode=log]
            
            \addplot+[scatter, only marks, scatter src=y] table[x  = beta, y  = val_decorr, col sep = semicolon] {data/search-mlpvae-decov-2D-norm.csv};
        \end{axis}		
    \end{tikzpicture}
    \begin{tikzpicture}
        \begin{axis}[height = 0.18 \textheight, width = 0.32\linewidth, xlabel={Beta}, ylabel={MAE}, grid style=dashed, legend pos = north east, legend style={font=\scriptsize}, tick label style={font=\scriptsize}, xmode=log]
            
            \addplot+[scatter, only marks, scatter src=y] table[x  = beta, y  = val_mae, col sep = semicolon] {data/search-mlpvae-decov-2D-norm.csv};
        \end{axis}		
    \end{tikzpicture}

    \begin{tikzpicture}
        \begin{axis}[height = 0.18 \textheight, width = 0.32\linewidth, xlabel={Gamma}, ylabel={KL Latent}, grid style=dashed, legend pos = north east, legend style={font=\scriptsize}, tick label style={font=\scriptsize}, xmode=log]
            
            \addplot+[scatter, only marks, scatter src=y] table[x  = gamma, y  = val_kl_latent, col sep = semicolon] {data/search-mlpvae-decov-2D-norm.csv};
        \end{axis}		
    \end{tikzpicture}
    \begin{tikzpicture}
        \begin{axis}[height = 0.18 \textheight, width = 0.32\linewidth, xlabel={Gamma}, ylabel={Decorrelation}, grid style=dashed, legend pos = north east, legend style={font=\scriptsize}, tick label style={font=\scriptsize}, xmode=log]
            
            \addplot+[scatter, only marks, scatter src=y] table[x  = gamma, y  = val_decorr, col sep = semicolon] {data/search-mlpvae-decov-2D-norm.csv};
        \end{axis}		
    \end{tikzpicture}
    \begin{tikzpicture}
        \begin{axis}[height = 0.18 \textheight, width = 0.32\linewidth, xlabel={Gamma}, ylabel={MAE}, grid style=dashed, legend pos = north east, legend style={font=\scriptsize}, tick label style={font=\scriptsize}, xmode=log]
            
            \addplot+[scatter, only marks, scatter src=y] table[x  = gamma, y  = val_mae, col sep = semicolon] {data/search-mlpvae-decov-2D-norm.csv};
        \end{axis}		
    \end{tikzpicture}
    \caption{Comparison of several disentangling and reconstruction metrics for the 2D case with $\beta-$Decorr VAE}
    \label{plots_betadecorrvae2D}
\end{figure}
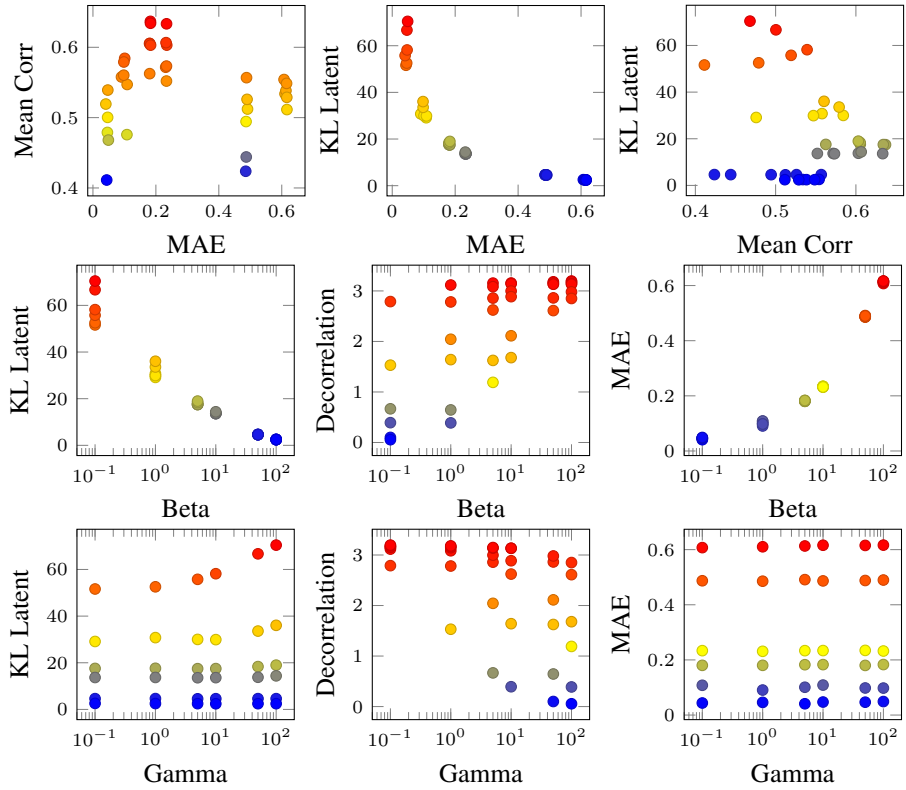

\clearpage
\FloatBarrier
\section{Selected Trajectory Plots}
\label{trajectories_sec}
In this section we present a selection of trajectories generated by different disentangling VAE setups. For this purpose we generate latent feature values for the $i$-th index as:

\begin{eqnarray}
    f_i = \left[ x_k = \begin{cases}
        0 & \text{if } i \neq k\\
        v & \text{else }i == k
    \end{cases} \right]_k
\end{eqnarray}

Latent features have 15 elements, where all of them are zeros, except for the $i$-th index which has a feature values $v$, that are generated by dividing the range $[-3, 3]$ into $n = 10$ equally spaced values. This range corresponds to the approximate range that a standard normal distribution would cover with high probability ($P(-3 < z < 3) = 0.997$).

A forward pass of the decoder is made with each feature vector, and the trajectories plotted. Plots are presented in left to right order with increasing feature values (from $-3$ to $3$).

For each trajectory, we compute the following score in order to determine its variation. This is similar to the coefficient of variation, but we normalize with the maximum value in order to enable comparison across different trajectories:

\begin{equation}
    s(x) = \frac{\text{std}(x)}{\max(x)}
\end{equation}

Many learned disentangled features collapse to produce little variation on the output trajectory when varying the feature value. We produce trajectories with all features and display the ones with the highest value of $S(x)$. We show the top three features for each disentangling method and selection of hyper-parameters $\beta$ and $\gamma$ (where applicable).

Each plot presents the trajectory as a scatter plot, with the starting point in green, and the end point in red. 

\subsection{Trajectories for VAE}

\begin{figure}[h]
    \includegraphics[width=\textwidth]{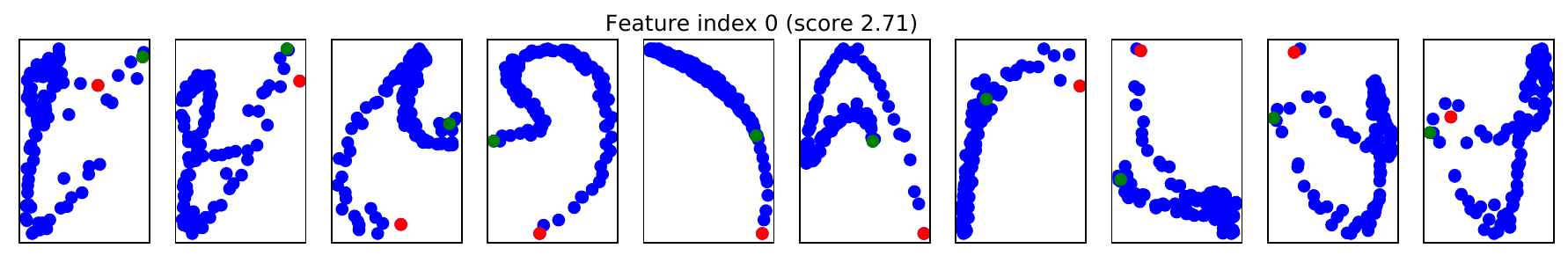}
    \includegraphics[width=\textwidth]{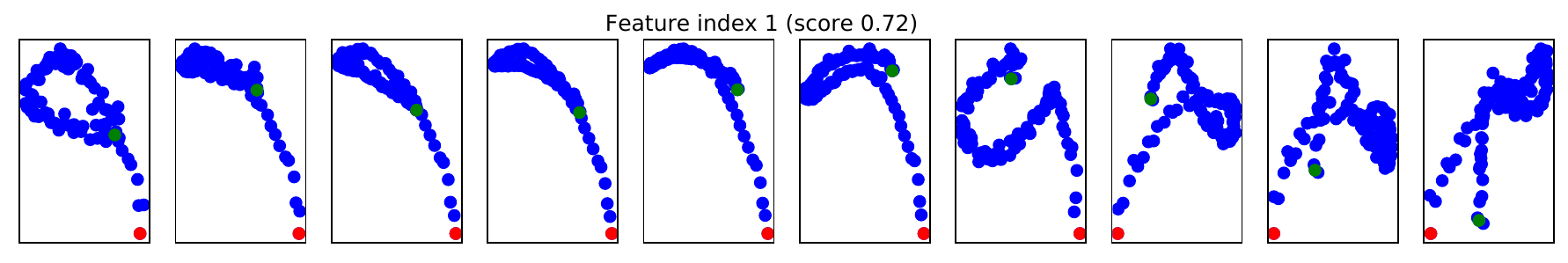}
    \includegraphics[width=\textwidth]{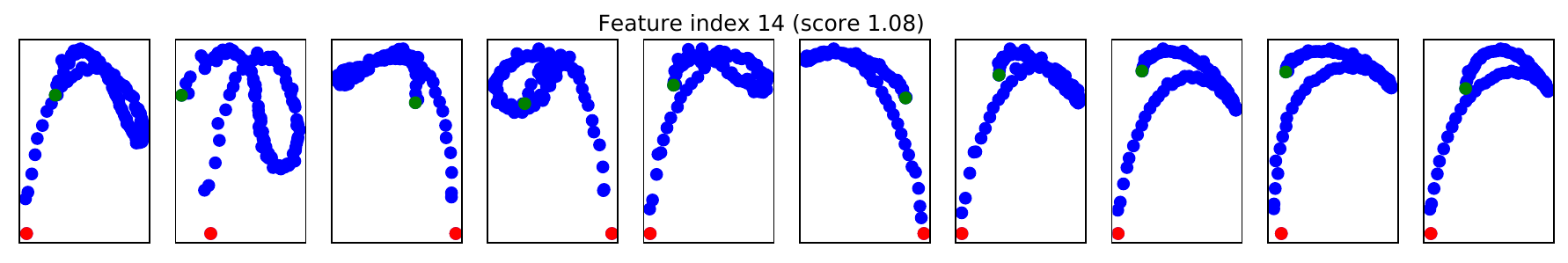}
    \caption{Trajectories generated by VAE. The green point is the trajectory start, and the red one is its end.}
\end{figure}

\FloatBarrier
\clearpage
\subsection{Trajectories for $\beta$-VAE}

\begin{figure}[h]
    \includegraphics[width=\textwidth]{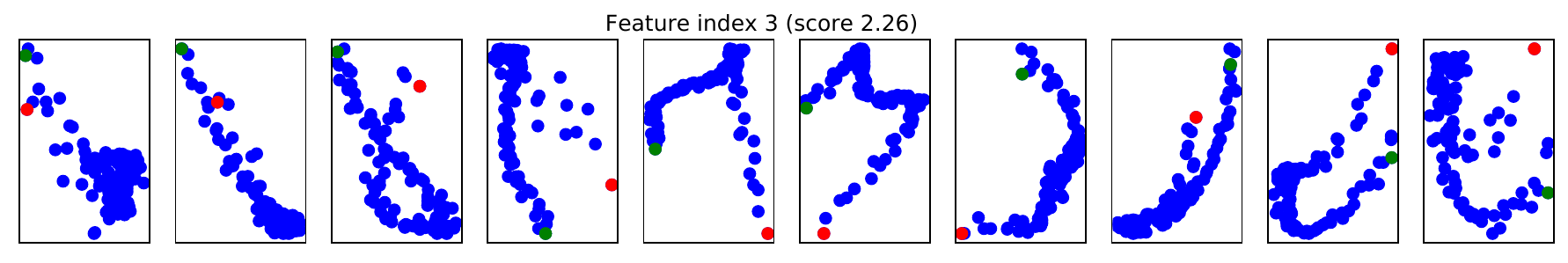}
    \includegraphics[width=\textwidth]{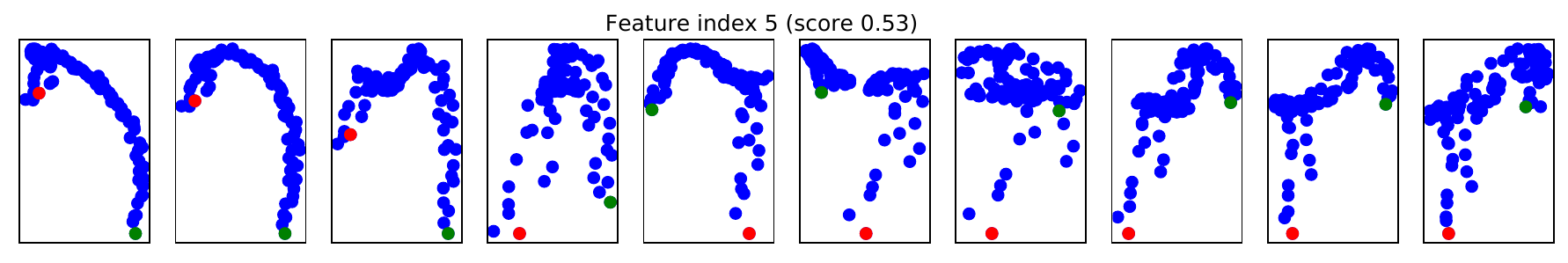}
    \includegraphics[width=\textwidth]{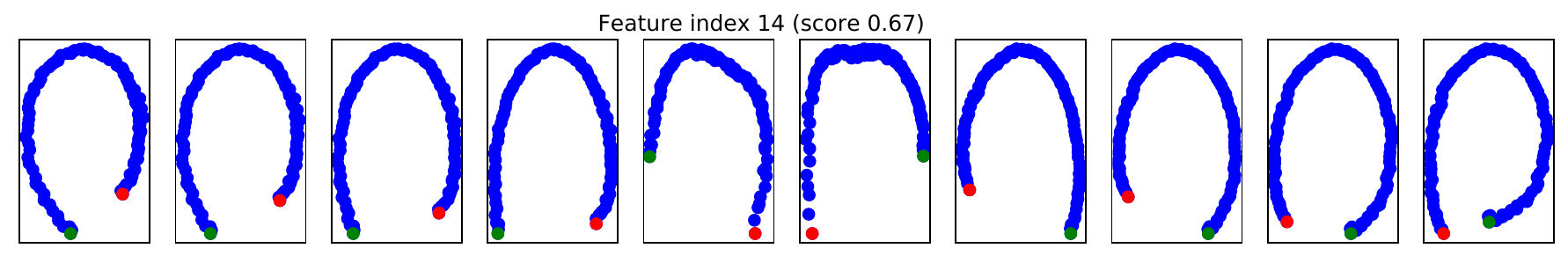}
    \caption{Trajectories generated by $\beta$-VAE with $\beta = 5.0$}
\end{figure}

\begin{figure}[h]
    \includegraphics[width=\textwidth]{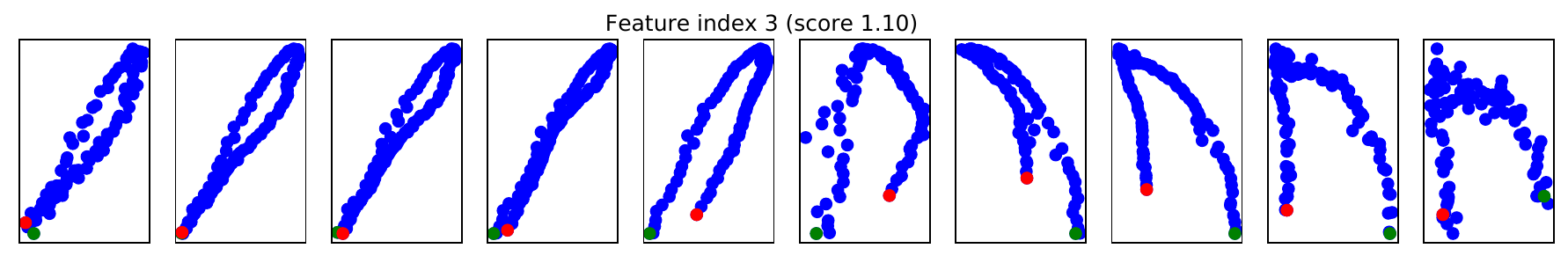}
    \includegraphics[width=\textwidth]{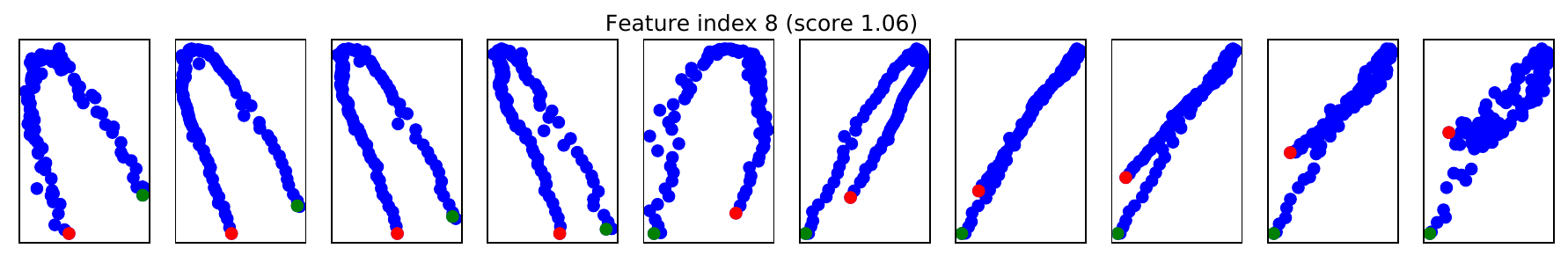}
    \includegraphics[width=\textwidth]{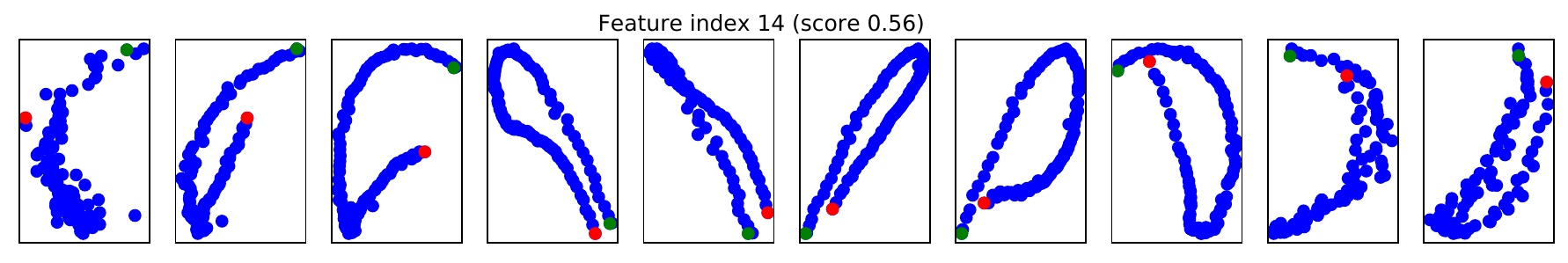}
    \caption{Trajectories generated by $\beta$-VAE with $\beta = 100.0$}
\end{figure}

\FloatBarrier
\clearpage
\subsection{Trajectories for Decorr VAE}

\begin{figure}[h]
    \includegraphics[width=\textwidth]{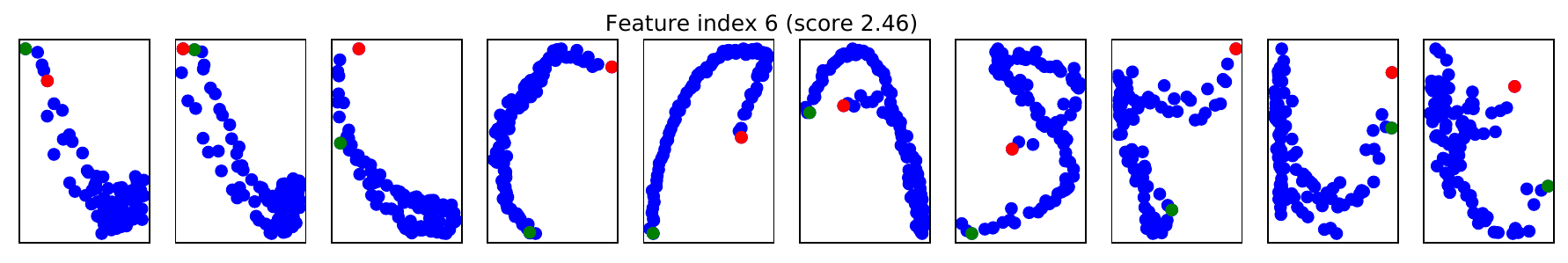}
    \includegraphics[width=\textwidth]{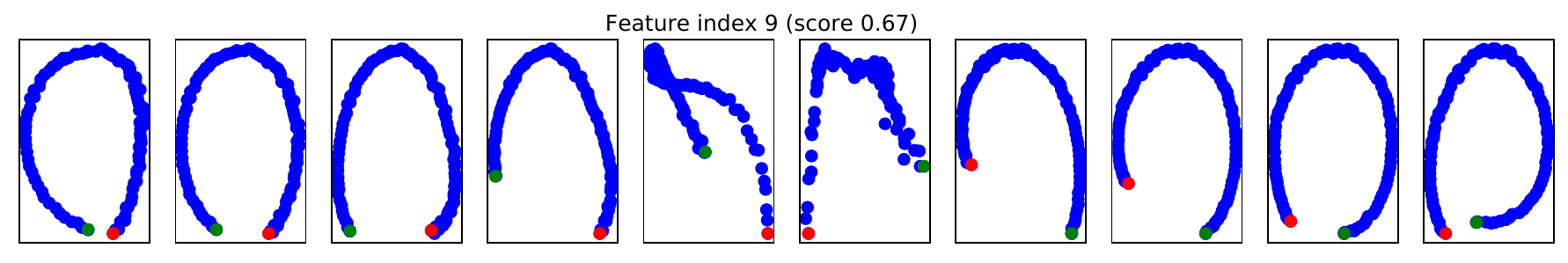}
    \includegraphics[width=\textwidth]{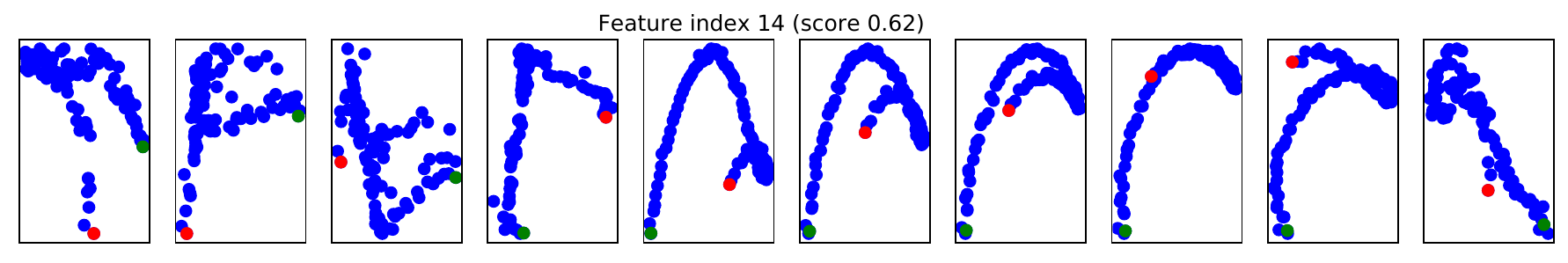}
    \caption{Trajectories generated by Decorr VAE with $\gamma = 0.1$}
\end{figure}

\begin{figure}[h]
    \includegraphics[width=\textwidth]{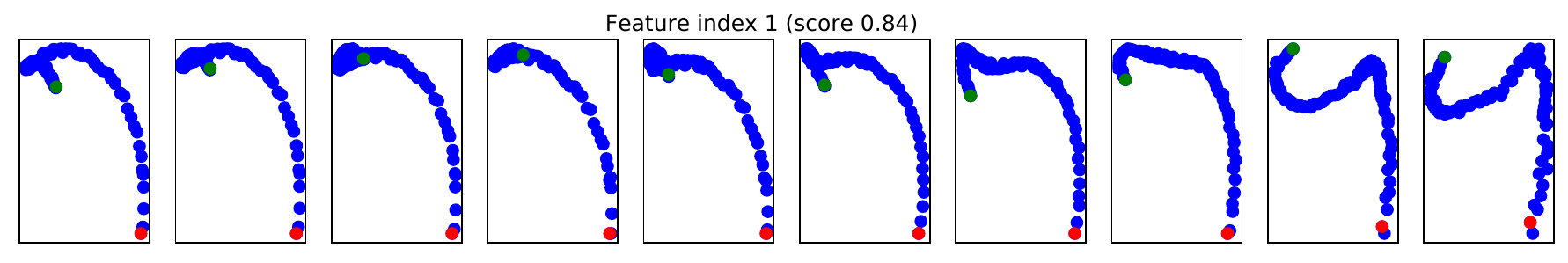}
    \includegraphics[width=\textwidth]{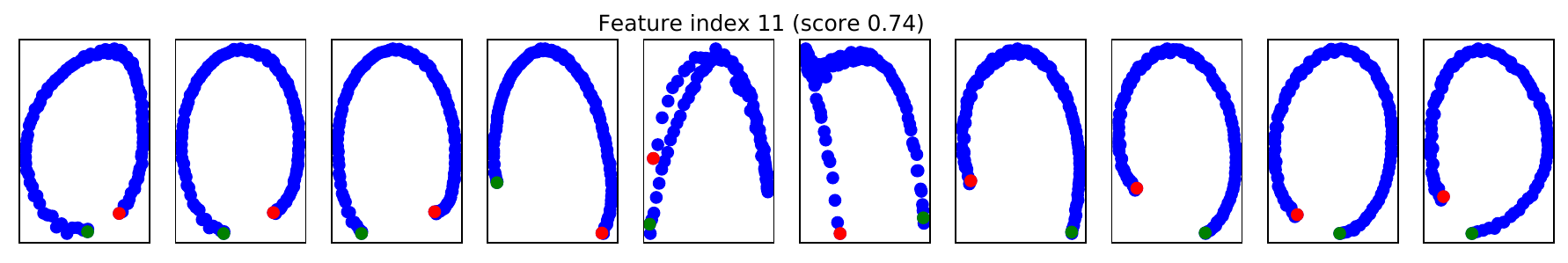}
    \includegraphics[width=\textwidth]{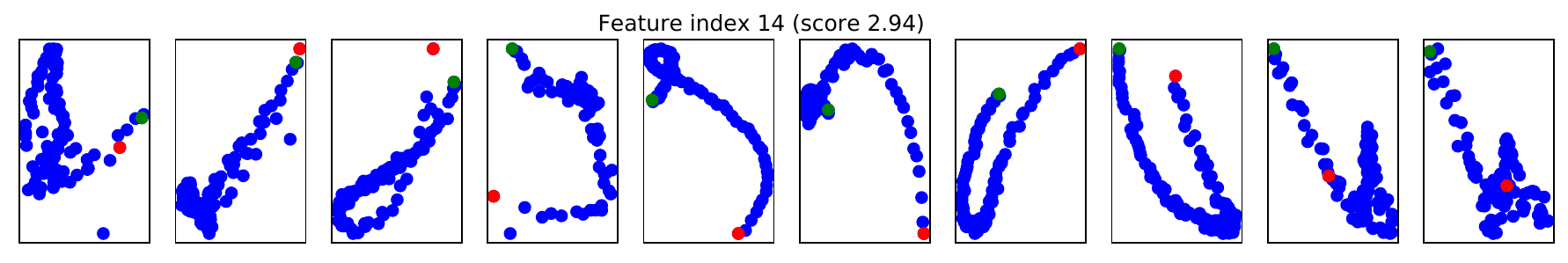}
    \caption{Trajectories generated by Decorr VAE with $\gamma = 5.0$}
\end{figure}

\begin{figure}[h]
    \includegraphics[width=\textwidth]{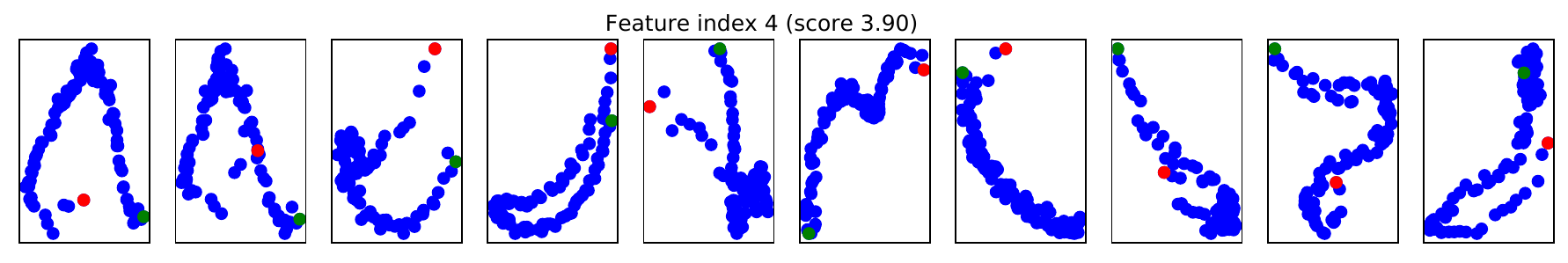}
    \includegraphics[width=\textwidth]{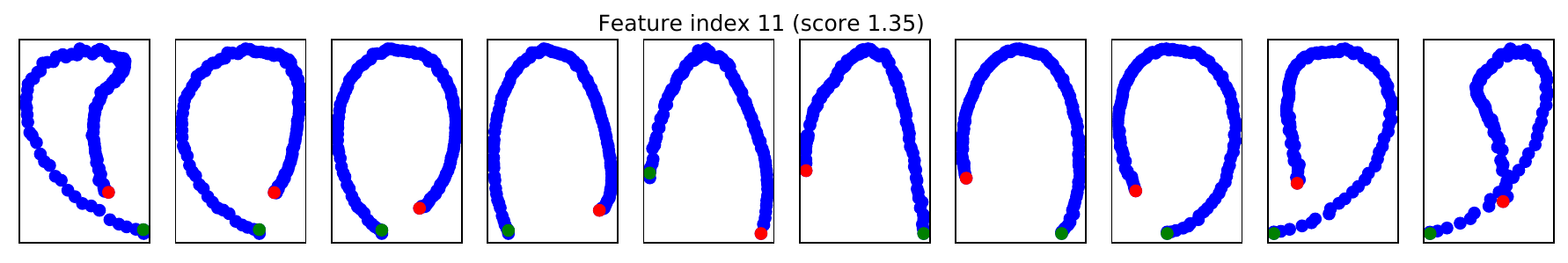}
    \includegraphics[width=\textwidth]{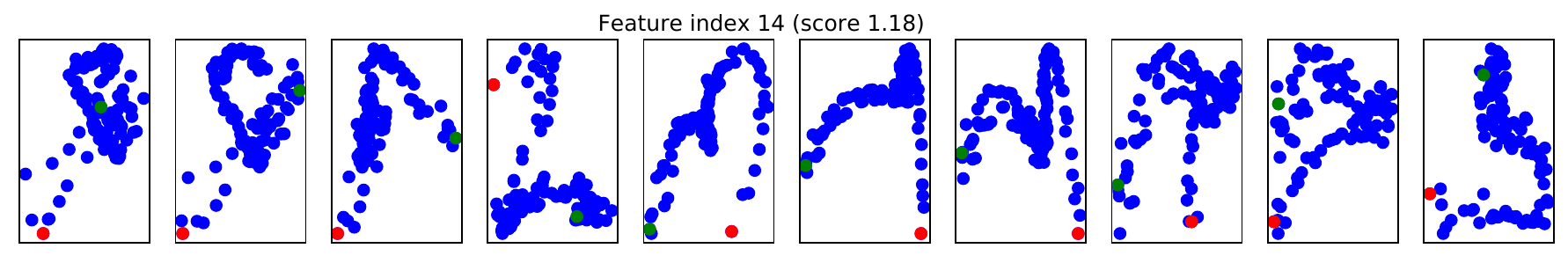}
    \caption{Trajectories generated by Decorr VAE with $\gamma = 100.0$}
\end{figure}

\FloatBarrier
\subsection{Trajectories for $\beta$-Decorr VAE}

\begin{figure}[h]
    \includegraphics[width=\textwidth]{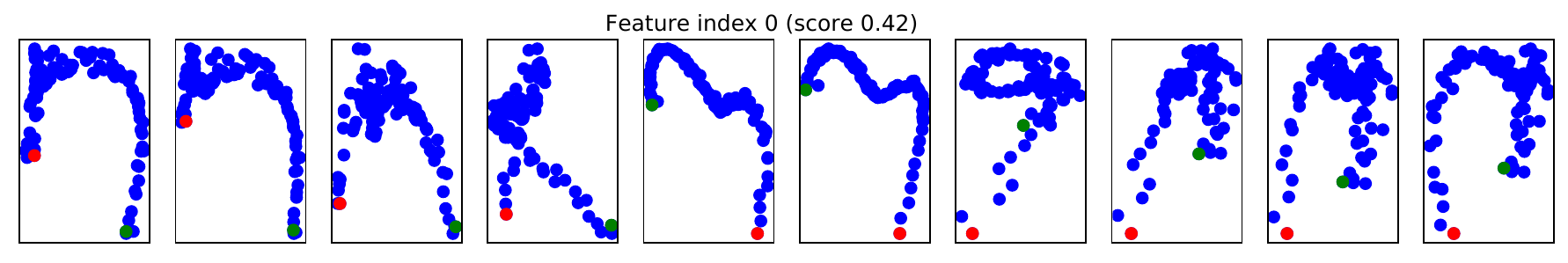}
    \includegraphics[width=\textwidth]{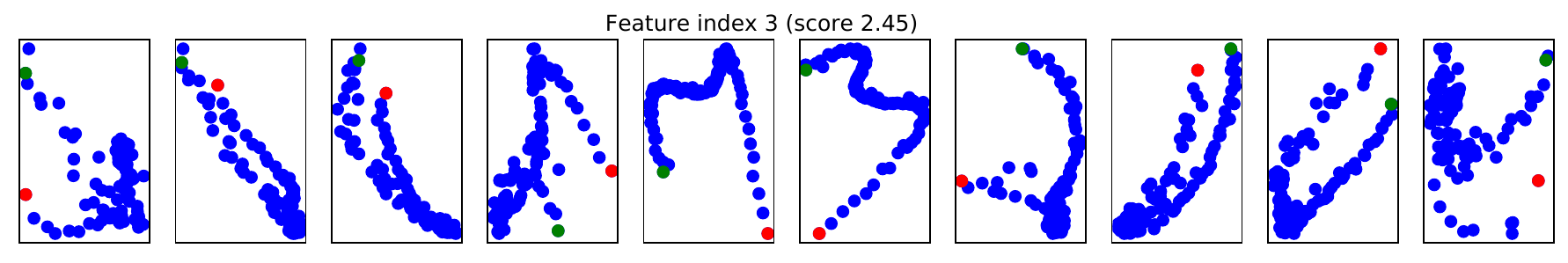}
    \includegraphics[width=\textwidth]{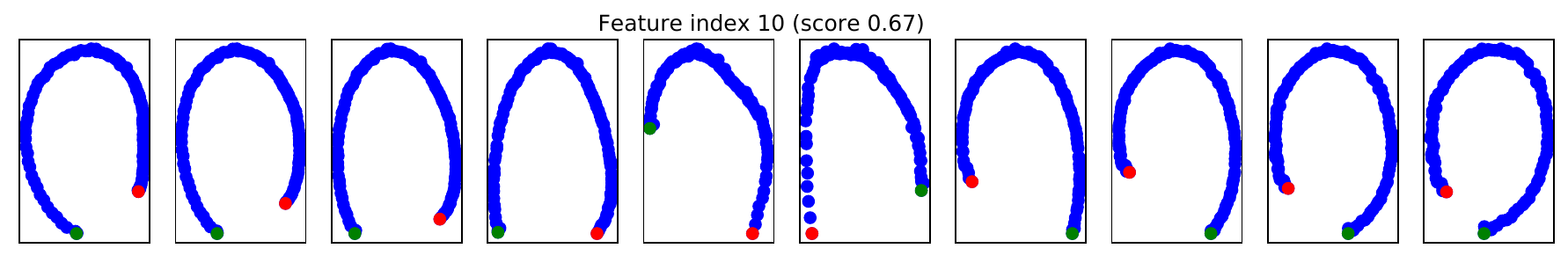}
    \caption{Trajectories generated by $\beta$-Decorr VAE with $\beta = 5.0$ and $\gamma = 5.0$}
\end{figure}

\begin{figure}[h]
    \includegraphics[width=\textwidth]{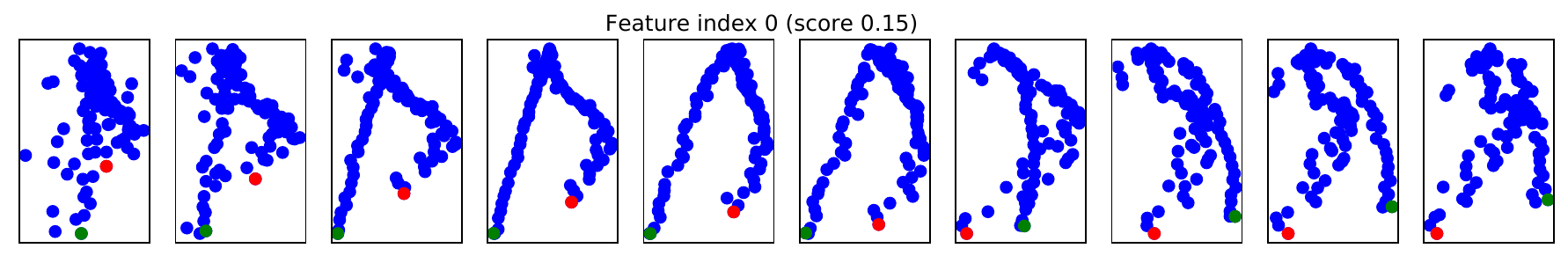}
    \includegraphics[width=\textwidth]{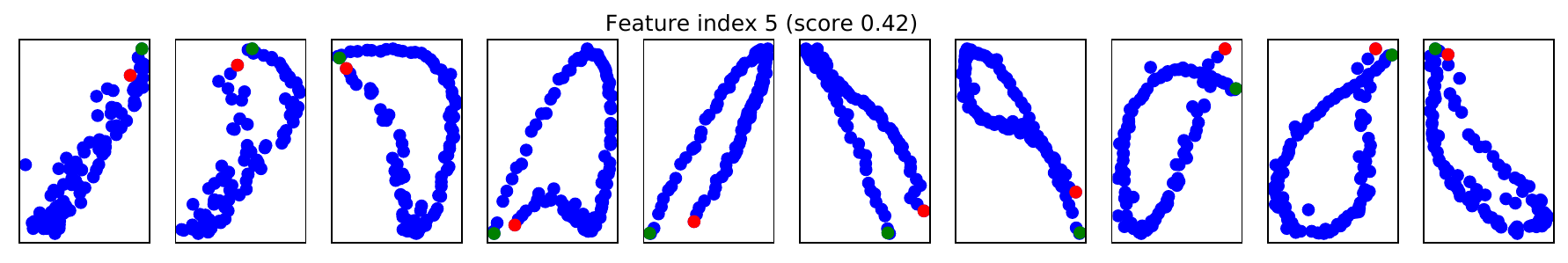}
    \includegraphics[width=\textwidth]{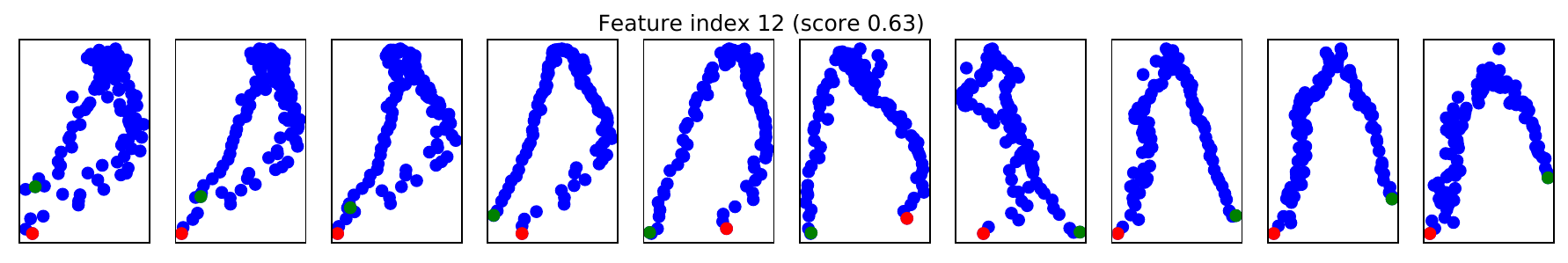}
    \caption{Trajectories generated by $\beta$-Decorr VAE with $\beta = 100.0$ and $\gamma = 10.0$}
\end{figure}

\begin{figure}[t]
    \includegraphics[width=\textwidth]{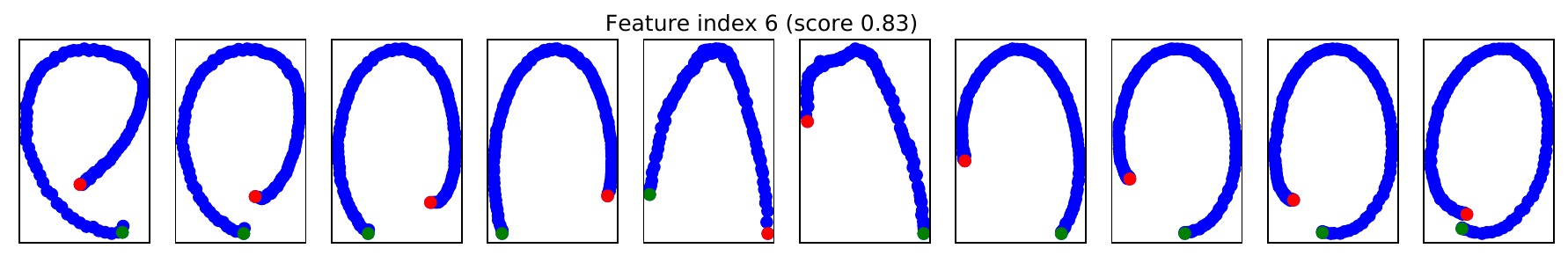}
    \includegraphics[width=\textwidth]{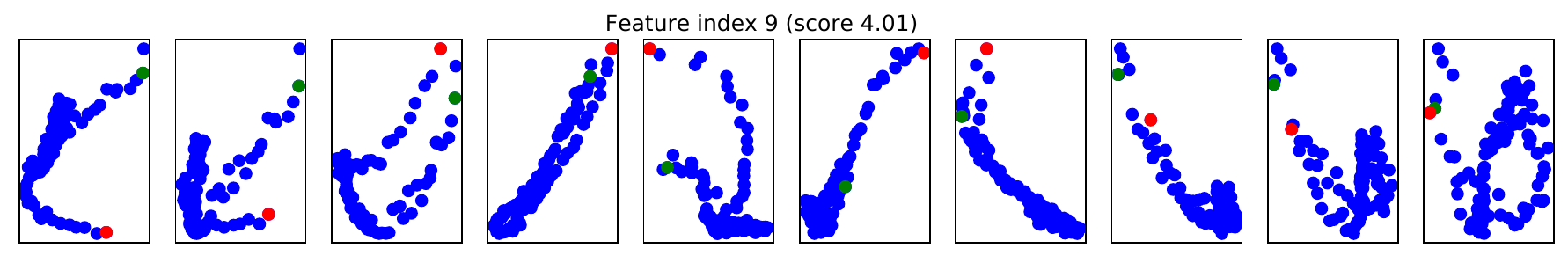}
    \includegraphics[width=\textwidth]{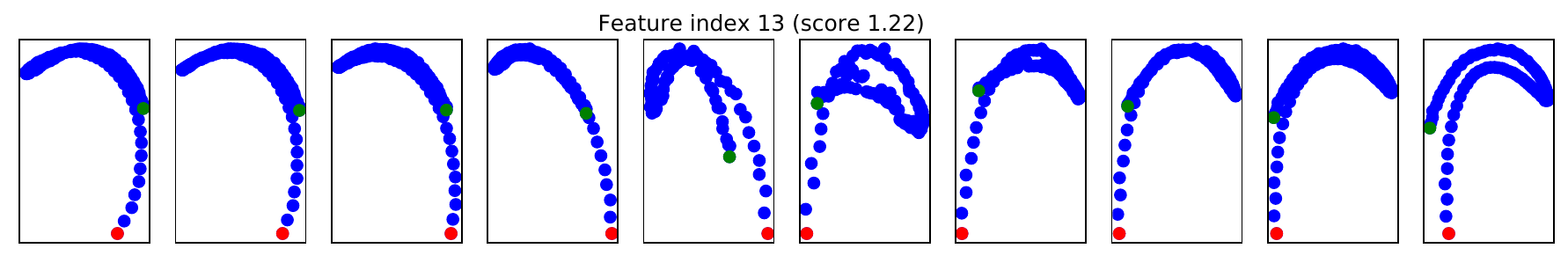}
    \caption{Trajectories generated by $\beta$-Decorr VAE with $\beta = 0.1$ and $\gamma = 5.0$}
\end{figure}

\end{document}